\documentclass[electronic]{vgtc}             

\graphicspath{{figures/}{pictures/}{images/}{./}} 

\usepackage{times}                     
\usepackage{amsmath}

\usepackage{tabu}                      
\usepackage{booktabs}                  
\usepackage{lipsum}                    
\usepackage{mwe}        
\usepackage{wrapfig}

\usepackage{mathptmx}                  
\usepackage{amsmath}
\usepackage{amssymb}

\RequirePackage{caption}                         
\RequirePackage{subfig}                          

\usepackage{booktabs} 
\usepackage{tabularx}
\usepackage{ragged2e}

\usepackage{multirow}
\usepackage[table]{xcolor}
\usepackage{graphicx}
\RequirePackage{color}                           
\RequirePackage[dvipsnames]{xcolor} 

\usepackage{svg}
\usepackage{colortbl}
\usepackage{pifont}
\newcommand{\cmark}{\ding{51}}%
\newcommand{\xmark}{\ding{55}}%

\usepackage{makecell}

\onlineid{1310}

\vgtccategory{Research}

\vgtcinsertpkg

\title{EPIC: \underline{Epi}polar-\underline{C}onsistent 360$^\circ$ Immersive Stereo Video Generation}

\author{Roy G. Biv\thanks{e-mail: roy.g.biv@aol.com}\\ %
        \scriptsize Starbucks Research %
\and Ed Grimley\thanks{e-mail: ed.grimley@aol.com}\\ %
     \scriptsize Grimley Widgets, Inc. %
\and Martha Stewart\thanks{e-mail: martha.stewart@marthastewart.com}\\ %
     \parbox{1.4in}{\scriptsize \centering Martha Stewart Enterprises \\ Microsoft Research}}

\author{Debabrata Mandal\\ %
        \scriptsize UNC Chapel Hill %
\and Dongdong Fu\\ %
     \scriptsize Dolby Laboratories %
\and Jonathon Miller\\ %
     \scriptsize Dolby Laboratories
\and William Villareal\\ %
     \scriptsize Dolby Laboratories
    \and Xi Peng\\ %
     \scriptsize UNC Chapel Hill
     \and Praneeth Chakravarthula\\ %
     \scriptsize UNC Chapel Hill}

\teaser{
  \centering
  \includegraphics[width=\linewidth]{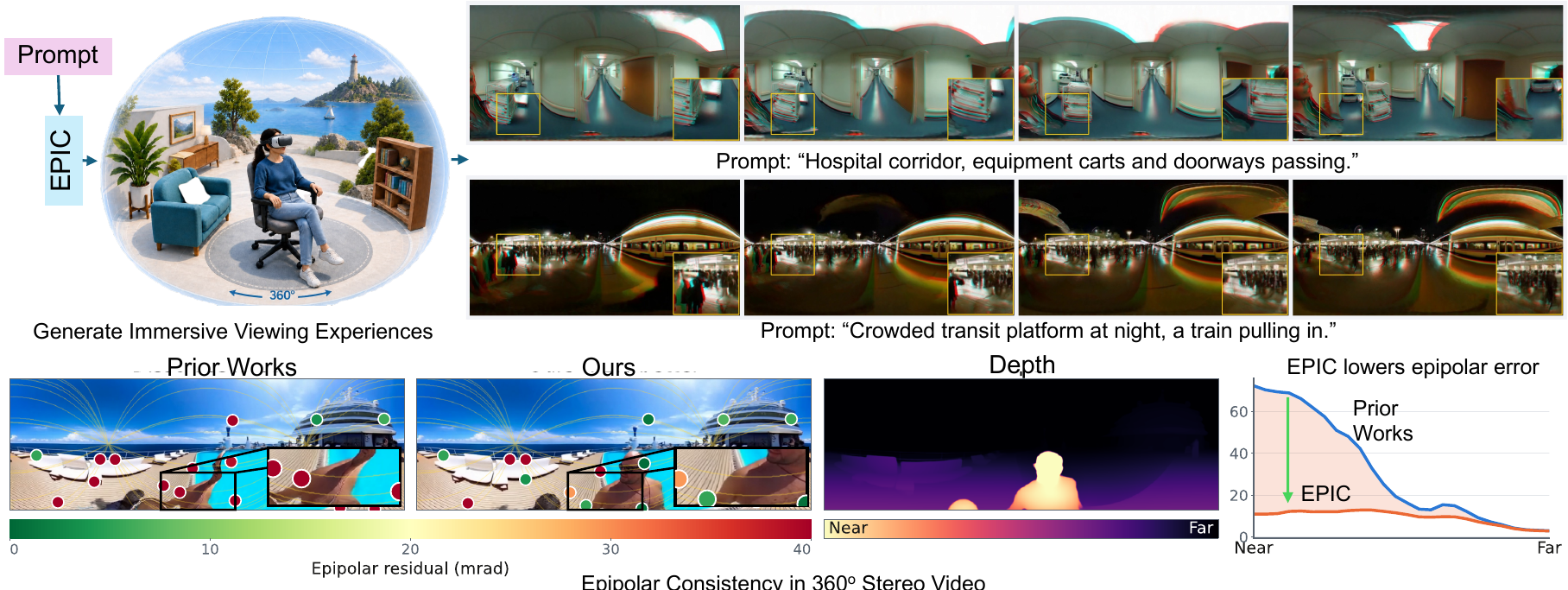}
  \caption{\textit{Teaser.} We present a panoramic stereo video refinement pipeline to create an immersive viewing experience directly from text \textit{or} image prompts. Our stereo generation pipeline works across a diverse range of scene prompt descriptions such as both indoors and outdoors. Prior works such as \textit{DissolveStereo} \cite{dissolvestereo} can only provide soft geometric supervision which leads to epipolar inconsistencies in the generated right views while \textit{ours} learns to produce lesser stereo errors through a novel approach.}
  \label{fig:teaser}
}

\abstract{
Immersive displays can enable rich and diverse virtual experiences.
Manually authoring every possible experience to realize this potential, however, is prohibitively expensive, difficult to scale, and impractical. 
Generative AI models could remove this bottleneck, but today’s models are built for conventional displays and cannot generate the high-resolution, stereoscopic $360^\circ$ content required for immersive viewing. 
Further, temporal and stereo inconsistencies that may be tolerable on conventional displays can become highly disruptive when viewed through an immersive headset. 

Here, we address this gap with a zero-shot generative pipeline that extends existing video diffusion models into 4K stereoscopic $360^\circ$ videos. 
Inspired from binocular vision and depth perception, we develop an epipolar-aware $360^\circ$ image matching metric that captures the temporal and stereo geometric inconsistencies across views.
We then use this metric as a preference signal for direct preference optimization with limited training data.
Our work enables $360^\circ$ stereo video generation and provides a scalable path for bringing generative content to immersive displays, allowing diverse mixed reality experiences on demand.

} 


\begin{document}


\firstsection{Introduction}



\maketitle

Immersive displays can transport viewers to places and experiences beyond their physical surroundings, from exploring remote environments to training in realistic simulations.
Realizing this potential, however, requires creating immersive content for every new experience, which is expensive and difficult to scale.
Capturing stereoscopic $360^\circ$ video requires complex multi-camera rigs~\cite{nayar1997catadioptric, stereo_capture}, careful stitching of captured images, and, most importantly, physical access to the scene being recorded.
Manually creating virtual environments requires artists to model, texture, and animate each scene.
In addition, neither approach can easily create experiences that cannot be staged or captured, such as dangerous, rare, or imagined environments.

Generative video models could remove this bottleneck by creating scenes directly from text or
  image prompts, enabling personalized experiences such as exposure therapy for
  phobias~\cite{carl2019vret}, vestibular and spatial
  rehabilitation~\cite{heffernan2021vestibular}, immersive skills
  training~\cite{humm2022vrsurgery}, and classroom education~\cite{coban2022learning}.
  However, today's models are primarily designed for conventional displays and cannot directly
  generate the high-resolution, stereoscopic $360^\circ$ video required for immersive viewing.
  Naively extending existing video generators to immersive displays introduces several
  challenges, such as incorrect geometry, that prior work does not fully address.
  While stereo generation~\cite{stereocrafter, dissolvestereo} and panoramic $360^\circ$
  generation~\cite{panowan, cubecomposer} have each been studied in isolation, their intersection
  remains underexplored.

The key obstacle to achieving immersive generative video is the lack of data. 
High-quality panoramic stereo video is scarce and expensive to capture, prohibiting the use of supervised training methods. 
Zero-shot methods such as DissolveStereo \cite{dissolvestereo} and StereoCrafter \cite{stereocrafter} avoid this requirement by diffusion inpainting and have been used for perspective stereo video.
However, these methods rely on soft conditioning guidance and do not explicitly enforce the correct geometry between the left and right eye views (see \cref{fig:teaser}). As a result, errors that may be tolerable on a conventional display become especially disruptive during binocular viewing.
Unfortunately, existing metrics also rely on perspective image matching~\cite{dissolvestereo}, which does not account for the spherical geometry of $360^\circ$ immersive visuals or the strong distortion of equirectangular projected (ERP) images near the poles.
We therefore need both a way to \textit{generate} immersive stereo video without large-scale stereo training data, and a way to \textit{measure and correct} the geometric errors that affect immersive viewing.

We address these challenges with a generative pipeline that extends existing video diffusion models to high-resolution stereoscopic $360^\circ$ video. 
We first introduce a zero-shot stereo adaptation procedure that allows existing video models to generate text- or image-conditioned stereo pairs without requiring large-scale training data.
However, the generated videos may contain incorrect disparities and other geometric inconsistencies between binocular views.
To identify these errors, we introduce a novel Panoramic Epipolar Geometry Score (PEGS), a ranking-based metric that measures stereo consistency directly on the $360^\circ$ viewing sphere.
Unlike perspective image matching, it accounts for spherical epipolar geometry. 
In addition, PEGS is pose-free and robust to independently moving objects, allowing it to evaluate the dynamic content produced by generative video models.
%
%
We use the PEGS score to rank generated panoramic stereo videos according to their geometric consistency and construct $(\textsc{preferred}, \textsc{dispreferred})$ pairs.
We then use these pairs to optimize the generator through direct preference optimization \cite{rafailov2023}. This process turns PEGS, a non-differentiable geometric metric, into a useful training signal and reduces the artifacts left by zero-shot stereo adaptation, without requiring ground-truth panoramic stereo video.

The current video generators are limited to 480p or 720p~\cite{wan}, well below the resolution needed for immersive viewing.
Therefore, we employ a stereoscopic panoramic upsampling stage that produces 4K output. 
We also demonstrate a stereo baseline adjustment procedure that controls the disparity between generated views, with implications for comfortable headset viewing.
Together, these components provide a scalable path to high-resolution stereoscopic $360^\circ$ generation for immersive displays.

In summary, this work makes the following contributions:

\begin{itemize}
    \item An immersive video generation pipeline, the first to the best of our knowledge, for generating 4K stereoscopic $360^\circ$ content without requiring large-scale immersive stereo training data.
    \item A pose-free epipolar consistency metric, PEGS, formulated directly on the $360^\circ$ viewing sphere, capturing geometric inconsistencies relevant to stereoscopic immersive viewing.
    \item A direct preference optimization method for correcting geometric artifacts within generated panoramic stereo videos without requiring ground-truth data.
    \item A stereoscopic upsampling and baseline adjustment method for bringing generated immersive content to headset-ready quality.
\end{itemize}

\section{Related Works}

Panoramic stereo captures the world as an egocentric observer experiences it,
  overcoming two complementary limitations: the restricted field of view of a
  pinhole camera, and the absence of depth in a monocular one. Early systems
  pursued each half of the problem separately, through catadioptric and
  multi-camera arrangements for panoramic capture~\cite{nayar1997catadioptric} and
  conventional binocular rigs for stereo~\cite{stereo_capture}, before converging
  on omnidirectional stereo, which synthesizes each stereo output view from a rotating left-right camera arm so that disparity is
  available in every viewing
  direction~\cite{peleg2001omnistereo}. The resulting optical setups are cumbersome to use. Dual
  fisheye and multi-rig designs demand precise calibration, and their stitched
  output carries seam artifacts and parallax errors wherever the assumed geometry
  breaks down~\cite{ho2017dualfisheye, xiang2018compact}; and the interocular
  baseline is fixed at capture time, leaving no way to re-target the content for a
  different display or viewer afterwards. Generative video models sidestep this
  problem entirely, producing realistic footage from text or images and reaching
  scenes that no rig could practically be built to
  film~\cite{video_gen_survey}. An ecosystem of model variants also allows arbitrary video resolution scaling and high dynamic range
  capabilities. This paper explores stereo generation for 360$^\circ$ videos through a fast and flexible adaptation pipeline.



\newcommand{\greencheck}{{\color{Green}\cmark}\xspace}
\newcommand{\green}{\cellcolor{Green!12.5}\greencheck}
\newcommand{\yellowcheck}{{\color{YellowOrange}(\cmark)}\xspace}
\newcommand{\yellow}{\cellcolor{YellowOrange!12.5}\yellowcheck}
\newcommand{\redcheck}{{\color{red}\xmark}\xspace}
\newcommand{\red}{\cellcolor{red!12.5}\redcheck}
\newcommand{\graycheck}{{\color{black}\textendash}\xspace}
\newcommand{\gray}{\cellcolor{gray!30.5}\graycheck}
\newcommand{\lock}{\faLock\xspace}
\newcommand{\unlock}{\textcolor{NavyBlue}{\faUnlock}\xspace}
\newcommand{\redstar}{{\color{red}{*}}\xspace}
\newcommand{\deb}[1]{\textcolor{Turquoise}{#1}}
\newcommand{\pc}[1]{\textcolor{blue}{#1}}
\definecolor{bestgray}{gray}{0.65}   
\definecolor{secondgray}{gray}{0.80} 
\definecolor{thirdgray}{gray}{0.92}  

\newcommand{\best}[1]{\cellcolor{bestgray}\textbf{#1}}
\newcommand{\second}[1]{\cellcolor{secondgray}#1}
\newcommand{\third}[1]{\cellcolor{thirdgray}#1}


\begin{table}[!t]
      \setlength{\tabcolsep}{0em}
      \centering
      \footnotesize
      \caption{
          Comparison of related work on panoramic and stereo video generation. Each criterion is
          fully~\greencheck, partially~\yellowcheck, or not met~\redcheck. Prior work covers
          either the panoramic axis or the stereo axis, but not both, and geometric structure
          enters as depth conditioning rather than as an explicit epipolar constraint.
      }
      \begin{tabularx}{\linewidth}{m{0.26\linewidth}
                                   >{\raggedright\arraybackslash}X
                                   >{\raggedright\arraybackslash}X
                                   >{\raggedright\arraybackslash}X
                                   >{\raggedright\arraybackslash}X
                                   >{\raggedright\arraybackslash}X
                                   >{\raggedright\arraybackslash}X}
      \toprule
      &
      {\footnotesize Pano-\newline Wan \cite{panowan}} &
      {\footnotesize Dissolve-\newline Stereo \cite{dissolvestereo}} &
      {\footnotesize Stereo-\newline World \cite{stereoworld}} &
      {\footnotesize Stereo-\newline WM \cite{stereoworldmodel}} &
      {\footnotesize Pano-\newline World-X \cite{panoworldx}} &
      {\footnotesize \textbf{Ours}} \\
      \midrule
      \multicolumn{7}{l}{\textbf{Output domain}}\\
      \midrule
      $360^{\circ}$ panoramic &
      \green & \red & \red & \red & \green & \green \\
      Stereo &
      \red & \green & \green & \green & \red & \green \\
      \midrule
      \multicolumn{7}{l}{\textbf{Geometry}}\\
      \midrule
      Depth-guided &
      \red & \green & \green & \green & \yellow & \green \\
      Epipolar constraint &
      \red & \red & \yellow & \yellow & \red & \green \\
      Camera control &
      \red & \red & \red & \green & \green & \green \\
      \midrule
      Model-agnostic &
      \green & \green & \yellow & \red & \red & \green \\
      \bottomrule
  \end{tabularx}
      \label{tab:related_work}
  \end{table}

\paragraph{Video generation models.} 
Contemporary video generators are latent diffusion transformers trained on
Internet-scale corpora~\cite{laion_aesthetics} by distribution matching, most commonly
through flow matching and its rectified variant~\cite{flow, rectified_flow}. Yet as the field has advanced rapidly toward more
capable systems such as Veo~3~\cite{veo3}, Wan-2.2~\cite{wan}, and
HunyuanVideo~\cite{hunyuan}, camera modality has received comparatively little
attention. Stereo generation has been studied in both trained~\cite{stereoworld}
and training-free~\cite{stereocrafter, dissolvestereo} regimes, but only for
pinhole cameras; panoramic video generators~\cite{cubecomposer, panowan} cover
the full sphere but render a single monocular viewpoint. This
objective asks the model to place mass where the data lies; it says nothing about
whether any individual sample obeys the physical regularities that every training
video necessarily obeys. The consequence is that models trained on billions of
3D-consistent videos still produce outputs that violate rigid object
geometry~\cite{vigor}, drift in scene depth and
scale~\cite{hollein2026world}, and fail the epipolar constraint between
frames~\cite{kupyn2025epipolar}. Because these properties are global to a sample
and cheap to verify but expensive to differentiate through, they are naturally
expressed as rewards rather than as likelihoods, and a line of recent work has
accordingly moved them into a post-training stage: preference optimization
against multi-view consistency~\cite{vigor} and against epipolar
error~\cite{kupyn2025epipolar} both recover structure that pretraining alone does
not supply. \Cref{tab:related_work} contrasts our method against the prior works within the video generation landscape. Our work rightfully addresses the combination of both panoramic and stereo content generation through rigid geometric consistency enforced through epipolar error.

\paragraph{Reinforcement Learning and Post-Training Alignment.} 
Post-training alignment steers a pretrained generator toward a desired property
by maximizing a reward while penalizing divergence from the reference model, a
formulation inherited from RLHF in language modeling~\cite{ouyang2022} and now
standard for diffusion and flow models. Approaches differ chiefly in how they
access the reward. Reward backpropagation methods such as DRaFT~\cite{draft} and
AlignProp~\cite{alignprop} differentiate through the sampling chain, which is
efficient but demands an end-to-end differentiable score. Policy gradient methods
including DDPO~\cite{ddpo}, DPOK~\cite{dpok}, and PPO-style variants accept
arbitrary non-differentiable rewards, but require many rollouts per update; for
video, where a rollout is a full denoising trajectory decoded to pixels, this is
prohibitive. Direct Preference Optimization~\cite{rafailov2023} and its
diffusion~\cite{wallace2024} and rectified-flow~\cite{liu2025videoreward}
extensions instead reduce alignment to a supervised loss over preference pairs,
requiring only a relative ordering between samples and never differentiating the
reward. The reward itself is usually learned, from aesthetic
classifiers~\cite{laion_aesthetics} or vision-language models trained on human
annotation~\cite{liu2025videoreward}. Such signals are costly to collect, encode
subjective judgments that need not track geometric correctness, and invite reward
hacking; models aligned this way have been shown to score well on the learned
metric while leaving epipolar error essentially
unchanged~\cite{kupyn2025epipolar}. Classical geometric criteria avoid these
failure modes by construction. The closest work to ours~\cite{kupyn2025epipolar}
adopts exactly this reasoning but computes its epipolar reward between frames
across time, where a single fundamental matrix holds only for static scenes under
camera motion. Our reward is measured between simultaneous stereo views, so one
essential matrix explains every correspondence regardless of object motion.

\section{Preliminary}
Here, we briefly discuss the key concepts used in our framework.
\subsection{Epipolar Constraint} 
\label{sec:epipolar_prelim} 

\begin{figure}[h]
    \centering
    \includegraphics[width=0.89\linewidth]{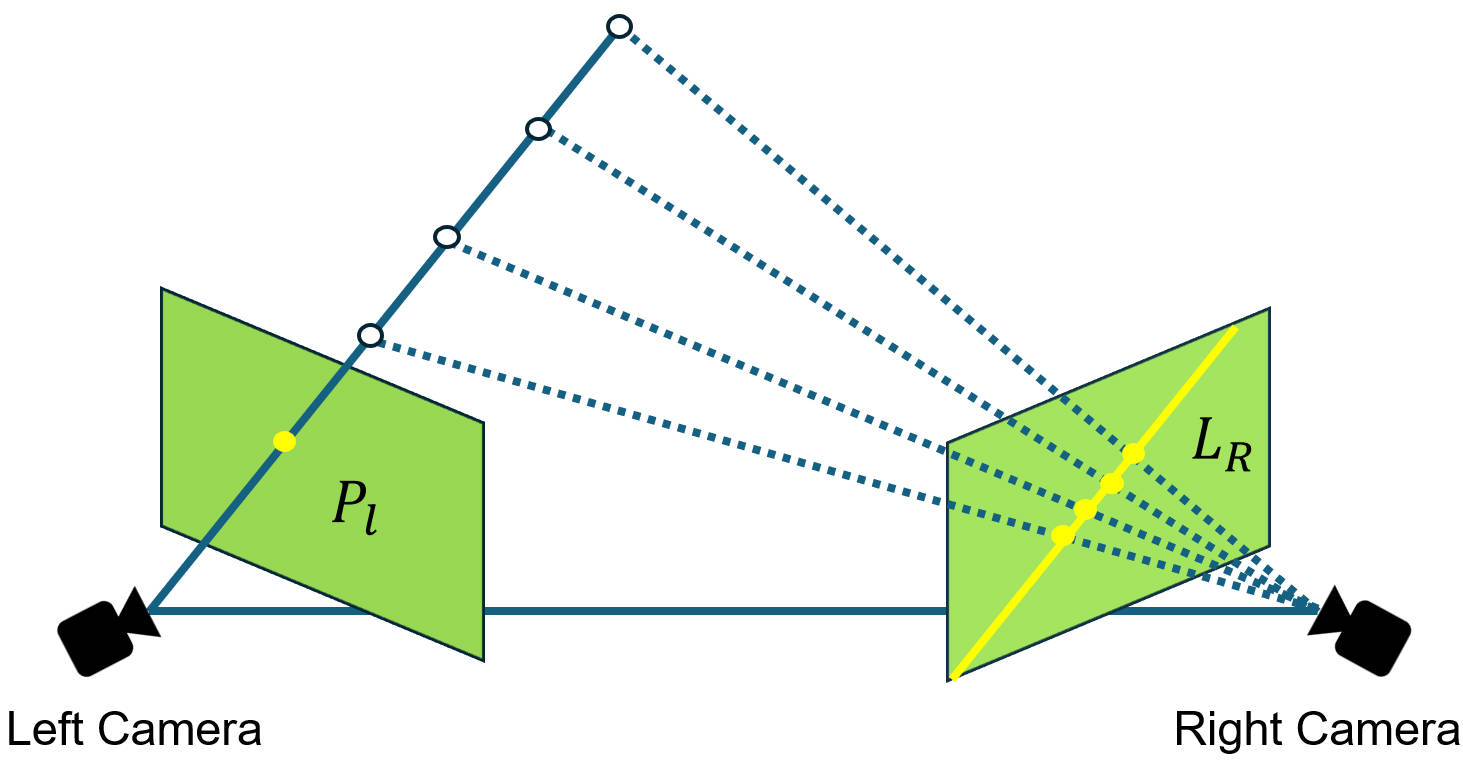}
    \caption{
        Illustration of the epipolar constraint. The observation on the left
        image plane $P_l$ determines a viewing ray, shown by the solid blue
        line. Because the depth is unknown, the observed 3D point may lie
        anywhere along this ray, as indicated by the white circles. When these
        possible 3D locations are projected into the right camera, their
        projections all lie on the yellow \textbf{epipolar} line $L_R$. Therefore, the
        point corresponding to the left observation must be found on $L_R$,
        rather than anywhere in the right image.
    }
    \label{fig:epipolar-constraint}
\end{figure}

As shown in Fig.~\ref{fig:epipolar-constraint}, an observation in the left
image determines the viewing direction $\mathbf{p}$ but not the depth of the
3D point. All possible 3D locations therefore lie along the same viewing ray.
This ray and the two camera centers define a plane called the
\emph{epipolar plane}. The corresponding viewing direction $\mathbf{p}'$ in
the right camera must also lie in this plane.

Consider a stereo pair whose two viewpoints are separated by a baseline
$\mathbf{t}$ (the vector between the two camera centers) and a relative
orientation $\mathbf{R}$ (the rotation taking one camera's frame into the
other's). A 3D scene point seen along the unit viewing direction $\mathbf{p}$
from the left camera and $\mathbf{p}'$ from the right must satisfy
%
\begin{equation}
    \mathbf{p}'^{\top} \mathbf{E}\, \mathbf{p} = 0,
    \qquad
    \mathbf{E} = [\mathbf{t}]_{\times} \mathbf{R},
    \label{eq:epipolar}
\end{equation}
where $\mathbf{E}$ encodes the two-view geometry (rotation $R$ and translation $t$ between the cameras) and $[\cdot]_{\times}$ denotes
the matrix form of the cross product. Equation~\eqref{eq:epipolar} states simply
that both viewing directions and the baseline are coplanar: they span the
\emph{epipolar plane}, and the correspondence to $\mathbf{p}$ must lie somewhere
in that plane. For a pinhole camera this plane cuts the image in a straight line. Rectification
  turns that line into a horizontal row ($L_R$ in \cref{fig:epipolar-constraint}), so perspective stereo reduces to a search
  along a scanline. For a panoramic camera the plane cuts the sphere in a great
  circle instead. Every such circle passes through the two points where the
  baseline pierces the sphere.



\subsection{Tangent Sampson Error} 
\begin{figure}[htbp]
    \centering
    \includegraphics[width=0.95\linewidth]{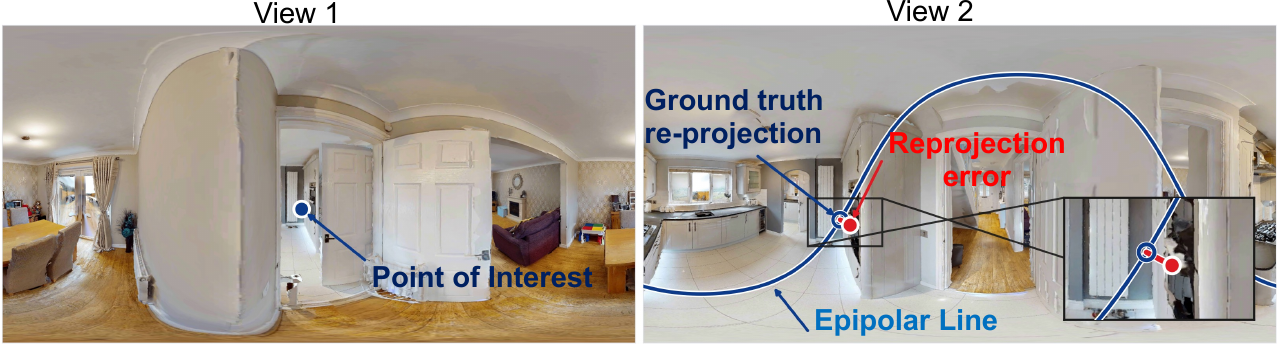}
    \caption{\textit{Tangent Sampson Error.} We compute the error by taking a point from one view and re-projecting it into another view and computing the reprojection error.}
    \label{fig:epipolar_error_viz}
\end{figure}

To turn the epipolar constraint introduced in \cref{sec:epipolar_prelim} into a usable error,
  one measures how far a correspondence is from satisfying it exactly. Or in other words, how far
  does the re-projected correspondence lie from the epipolar line in the other camera view as
  shown in \cref{fig:epipolar_error_viz}. For a pinhole pair, the classical Sampson
  error~\cite{hartley2004} measures exactly this,
  \begin{equation}
      E_{\mathrm{S}}^{2} =
      \frac{\left(\mathbf{p}_2^{\top} \mathbf{E}\, \mathbf{p}_1\right)^{2}}
           {\bigl\lVert [\mathbf{E}\,\mathbf{p}_1]_{1:2} \bigr\rVert^{2}
          + \bigl\lVert [\mathbf{E}^{\top}\mathbf{p}_2]_{1:2} \bigr\rVert^{2}},
      \label{eq:sampson}
  \end{equation}
  the constraint violation $\mathbf{x}_2^{\top}\mathbf{E}\,\mathbf{x}_1$ divided by how fast it
  changes when the image points move, with $[\,\cdot\,]_{1:2}$ taking the first two components.
  The denominator is where the pinhole assumption sits: it treats the two image coordinates as the
  free parameters of a measurement, which is false for a panorama.

  Terekhov and Larsson~\cite{terekhov2023tangent} keep the same ratio but replace the image points
  with unit viewing directions $\mathbf{d}_i = \pi^{-1}(\mathbf{p}_i)$, where $\pi$ is the camera
  projection, and the image derivatives with $J_i^{\dagger}$, the pseudoinverse of the Jacobian of
  $\pi$ at $\mathbf{d}_i$, giving the \emph{Tangent Sampson} error
  \begin{equation}
      E_{\mathrm{TS}}^{2} =
      \frac{\left(\mathbf{d}_2^{\top} \mathbf{E}\, \mathbf{d}_1\right)^{2}}
           {\lVert \mathbf{d}_2^{\top} \mathbf{E} J_1^{\dagger} \rVert^{2}
          + \lVert \mathbf{d}_1^{\top} \mathbf{E}^{\top} J_2^{\dagger} \rVert^{2}},
      \label{eq:tangent_sampson}
  \end{equation}
  which reduces to the classical Sampson error when $\pi$ is a pinhole projection.
Because $J_1$ and $J_2$ depend only on the measured points and not on
$\mathbf{E}$, they can be precomputed, making the residual nearly as cheap to
evaluate as the pinhole form; we refer the reader
to~\cite{terekhov2023tangent} for the derivation and for extensions handling
measurement covariances. For the equi-rectangular panoramic projection used here, $\pi$ has
a closed-form Jacobian, so Equation~\eqref{eq:tangent_sampson} is directly
applicable to our setting.

\begin{figure*}[htbp]
    \centering
    \includegraphics[width=0.95\textwidth]{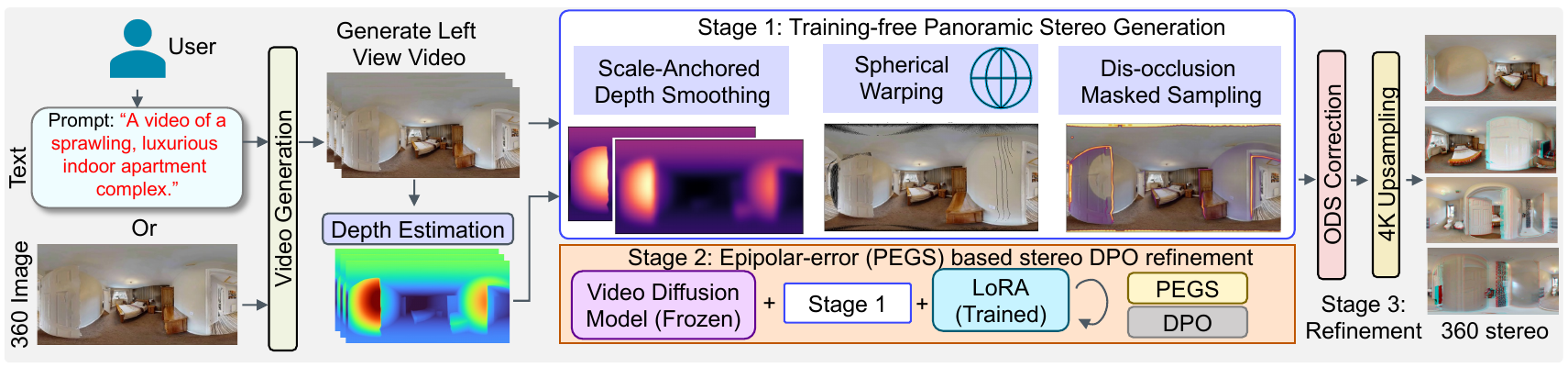}
    \caption{\textit{Overview.} We present a 360$^\circ$ stereo generation pipeline to convert user provided text or image prompts into headset viewable content. Our pipeline works in three main stages a training-free stereo approach to provide initial stereo estimates, a preference based finetuning approach without requiring expensive 360$^\circ$ stereo data collection and a final post-processing stage for immersive viewing.}
    \label{fig:overview}
\end{figure*}

\subsection{Flow Matching for Video Generation}
\label{sec:prelim_flow}

We build on video generators trained with rectified flow \cite{liu2023flow, polyak2024moviegen, jin2024pyramidal}, which learn how to transform a noise latent into a clean video latent by following a time-dependent vector field
.
Let $x_0 \sim q(x_0 \mid c)$ denote a clean video latent produced
by a temporal VAE (variational autoencoder), conditioned on $c$ (a text prompt and, for image-to-video,
a reference frame), and let $x_1 \sim \mathcal{N}(\mathbf{0}, \mathbf{I})$ be an independently sampled Gaussian-noise latent noise with the same dimensions as $x_0$. Rectified flow defines a linear interpolation
between the two,
\begin{equation}
    x_t = (1-t)\,x_0 + t\,x_1, \qquad t \in [0,1],
    \label{eq:rf_interp}
\end{equation}
where $t$ corresponds to the time steps along the straight path from noise to clean video latent, and the velocity $v = (x_t - x_0)/t$ refers to the directional flow of change in video latent space. Noticeably, the velocity $v$ is constant as $x_1 - x_0$, throughout $t \in [0, 1]$.
A transformer-based model $v_\theta(x_t, t, c)$ parameterized by $\theta$ is trained to predict the target velocity from the intermediate latent ($x_t$), time $t$, and condition $c$. The flow-matching objective is 
\begin{equation}
    \mathcal{L}_{\mathrm{FM}}(\theta) =
    \mathbb{E}_{t,x_0}
    \left\lVert v - v_\theta(x_t, t, c) \right\rVert^2 ,
    \label{eq:rf_objective}
\end{equation}
%
At inference, generation starts from a noise latent at $t=1$ and integrates the learned
  velocity field back to $t=0$; we write $p_\theta(x_0 \mid c)$ for the distribution of clean
  latents this produces. Separately, we abbreviate the per-sample velocity error inside
  \cref{eq:rf_objective} as
  $\mathcal{V}_\theta(x_t, t) = \lVert v - v_\theta(x_t, t, c)\rVert^2$. Both are needed in \cref{sec:method_dpo}: steering the generator is posed as moving
  $p_\theta$ toward stereo-consistent outputs while keeping it close to the pretrained model,
  and the objective that does so is written entirely in terms of $\mathcal{V}_\theta$.

\section{Method} Our proposed panoramic stereo generation and epipolar refinement pipeline is comprised of three main stages, as shown in \cref{fig:overview}. Here, we describe each stage in detail.

\begin{figure}[h]
    \centering
    \includegraphics[width=\linewidth]{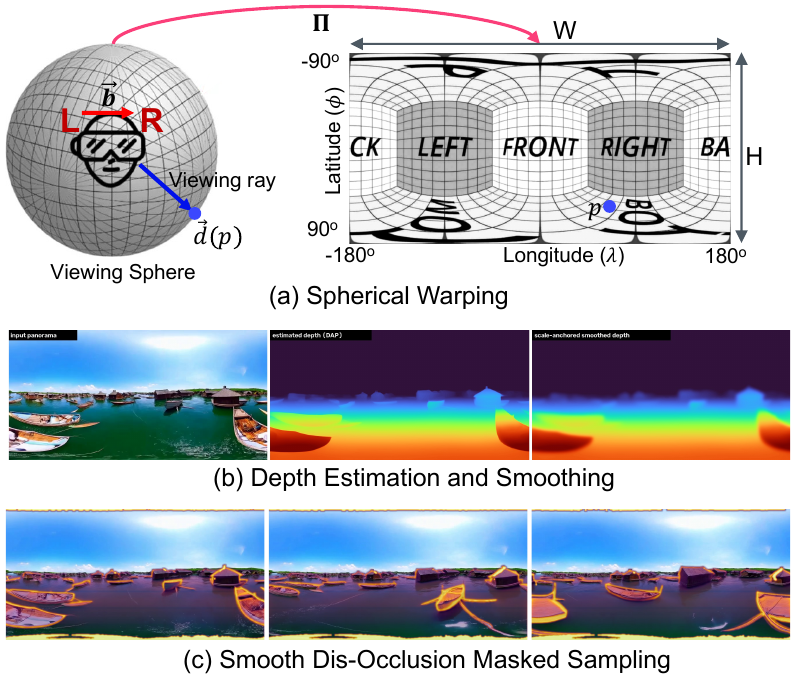}
    \caption{\textit{Stage 1: Zero-shot 360$^\circ$ Stereo Generation.} Our training-free stage can produce a large collection of 360$^\circ$ stereo videos without requiring any training data. We perform depth-wise spherical warping to get the right view from the generated left view and in-paint the occluded regions using the video diffusion model.}
    \label{fig:stage1}
\end{figure}

\subsection{Stage 1: Training-free Panoramic Stereo}
\label{sec:stage1}

  Video generation models are largely trained on monocular pinhole videos collected from the
  Internet such as WebVid~\cite{webvid}. Diffusion inpainting methods~\cite{sdedit} are used to
  fill in missing regions in generated content by sampling from the posterior of the learned
  prior, filling in physically plausible content. However, central camera models such as fisheye
  $180^\circ$ or panoramic $360^\circ$ do not follow the same image capture process as pinhole
  cameras. Hence, we first perform a zero-shot stereo generation pipeline
  (\cref{fig:stage1}.a) ~\cite{dissolvestereo} over a pre-trained panoramic
  generation model~\cite{panowan}. To accommodate the all-around viewing of panoramic cameras, we perform spherical
  warping from the left to the right view using estimated video depth. Let $\mathbf{p}=(u,v)$
  index the equirectangular grid and $\mathbf{d}(\mathbf{p})$ be the unit viewing ray of pixel
  $\mathbf{p}$,
  \begin{equation}
  \mathbf{d}(\mathbf{p}) =
  \begin{bmatrix}\cos\phi\sin\lambda \\ \sin\phi \\ -\cos\phi\cos\lambda\end{bmatrix},
  \quad
  \lambda = \tfrac{2\pi u}{W}-\pi,\;\; \phi = \tfrac{\pi}{2}-\tfrac{\pi v}{H},
  \end{equation}
  where $\lambda$ and $\phi$ are the longitude and latitude of $\mathbf{p}$, $H$ and $W$ are the height and width of generated image dimension. Let
  $\Pi:\mathbb{S}^2\rightarrow[0,W)\times[0,H)$ be the forward equirectangular projection. We
  extend it to any $3$D point by normalizing before projecting. For a rigid two-eye view with baseline $\mathbf{b}$ representing the direction from left to right eye, the
  right view then follows by pushing each ray out to its depth and re-projecting it from the left eye,
  \begin{equation}
  \hat{I}_R(\mathbf{p}) \;=\; I_L\!\Big(\Pi\big(\,\overline{D}(\mathbf{p})\,
  \mathbf{d}(\mathbf{p}) + \mathbf{b}\,\big)\Big),
  \qquad
  \overline{D} = \max\!\big(D,\; D_{\min}\big).
  \label{eq:warp}
  \end{equation}
  Here $D$ is the estimated video depth. We obtain it with a panoramic image depth estimation
  model~\cite{dap} and then smooth it temporally (\cref{fig:stage1}.b). 


  The warp cannot fill everything. Surfaces the right eye can see but the left eye never observed
  have no source pixel, leaving holes along depth edges. We mark these with a smooth binary mask (\cref{fig:stage1}.c) and let
  the generator synthesize inside it only, keeping the warped content everywhere else, so the two
  views agree by construction wherever the geometry already fixes them.

\subsection{Panoramic Epipolar Geometry Score}

  The zero-shot pipeline from \cref{sec:stage1} does not enforce correct geometry. The generated content may not follow the warped content closely about where the points in the observed 3D space may lie. Hence, we need a metric
  that can measure geometric dis-agreement directly, which appearance metrics simply cannot do, since two views
  can each look correct while disagreeing about the underlying 3D scene (\cref{fig:epipolar-constraint}).

\begin{figure}[h]
    \centering
    \includegraphics[width=\linewidth]{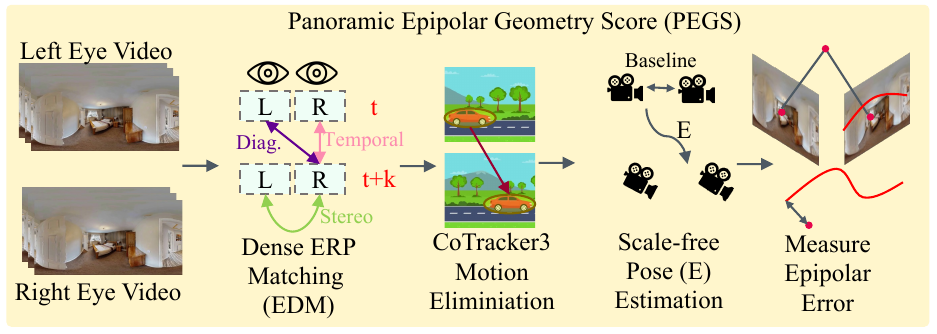}
    \caption{\textit{Stereo Scorer.} Our panoramic stereo scorer metric performs robust feature matching to find compatible point correspondences for epipolar error computation.}
    \label{fig:scorer}
\end{figure}

  We measure that disagreement as a violation of the epipolar constraint from
  \cref{sec:epipolar_prelim}. For two views of a rigid scene, every true correspondence must
  satisfy it, and we report the violation as an angle on the viewing sphere.
  \Cref{fig:scorer} shows the steps.

  We begin with two views and find dense correspondences using a matcher built for
  equirectangular images~\cite{edm}, keeping only confident matches. Each match gives a pair of
  unit viewing rays $\mathbf{d}_A$ and $\mathbf{d}_B$, one in each view. Moving objects break the
  rigid-scene assumption, so we track points across frames~\cite{cotracker} and discard those
  that move differently from the background scene such as \textit{cars} or \textit{people}. The surviving matches give the relative pose i.e. the rotation, and the direction the camera
  moved. Correspondences, however, cannot reveal how far the camera moved, and the metric does not need it either. It
  asks whether two viewing rays meet, which does not depend on how far apart the cameras are.  We then evaluate the Tangent Sampson error $\varepsilon$ of
  \cref{eq:tangent_sampson} on every match. It returns an angle in radians and is defined
  everywhere on the sphere, even in the poles where plain Sampson error is undefined \cite{terekhov2023tangent}.

  We repeat this for three pairs of views: the two eyes at the same time instant (t) ($L_t$ and $R_t$), one eye at two
  instants ($t$ and $t+k$), and one eye against the other eye at a later instant (\cref{fig:scorer}). The last
  pair is particularly important for rigid two-view geometry. A model can produce a plausible stereo pair and a plausible temporal pair without any coherent scene behind them, but it cannot satisfy the combined
  constraint unless such a scene exists.

  A score that can be gamed will be, so the metric reports whether it is valid instead of always returning a number. A right view copied from the left drives the metric score to zero while destroying the stereo effect, so we require real parallax to be present. A static camera makes the temporal pair vacuous. Too few surviving matches make any estimate unreliable. The reported
  score is
  \begin{equation}
  \mathrm{PEGS} \;=\; \frac{10^{3}}{\lvert\mathcal{T}\rvert}
  \sum_{\kappa\in\mathcal{T}}
  \operatorname*{median}_{(A,B)\in\mathcal{C}_{\kappa}}
  \varepsilon\big(\mathbf{d}_A,\mathbf{d}_B;\, E_{\kappa}\big),
  \label{eq:pegs}
  \end{equation}
  in milliradians, where $E_{\kappa}$ is the essential matrix for pair $\kappa$, $\mathcal{T}$
  collects the pairs that pass these checks, and $\mathcal{C}_{\kappa}$ the matches kept for pair
  $\kappa$. A pair that fails is dropped rather
  than set to zero, since treating it as zero would rank a degenerate clip best.

\subsection{Stage 2: Preference Optimization for Stereo}
\label{sec:method_dpo}

 Let $p_{\mathrm{ref}}$ be the distribution of generated videos from the pretrained video
  generative model~\cite{wan}, kept frozen. We want a model that produces stereo-consistent
  videos $p_\theta$ without losing the generative quality of $p_{\mathrm{ref}}$. The standard way
  to write this goal is
  \begin{equation}
      \max_{\theta}\;
      \mathbb{E}_{c \sim \mathcal{P},\, x_0 \sim p_\theta}
      \left[ r(x_0) \right]
      - \beta\, D_{\mathrm{KL}}\!\left( p_\theta \,\|\, p_{\mathrm{ref}} \right),
      \label{eq:rlhf}
  \end{equation}
  where $r(x_0)$ is a reward that is high for the property we want and $\mathcal{P}$ is the set of
  text conditionings we train on. The Kullback--Leibler divergence $D_{\mathrm{KL}}$ measures how far
  $p_\theta$ has moved from $p_{\mathrm{ref}}$, and $\beta$ sets how much movement is allowed.

  Two things stop us from optimizing \cref{eq:rlhf} directly. PEGS needs feature matching and
  two-view estimation on a decoded video, so it has no gradient with respect to $\theta$. It must
  also be computed freshly on samples from $p_\theta$ every training step, which would require an expensive inference pass through the model. Direct Preference Optimization~\cite{rafailov2023} removes both. The model that maximizes
  \cref{eq:rlhf} is known in closed form~\cite{peng2019awr,korbak2022,rafailov2023}, which
  simplifies the whole objective to asking only which of two samples is better. We still train
  the model, but never have to differentiate the reward. This suits PEGS, which ranks samples
  reliably under identical conditioning even though its absolute value does not compare across
  scenes~\cite{kupyn2025epipolar}.

\begin{figure}[h]
    \centering
    \includegraphics[width=\linewidth]{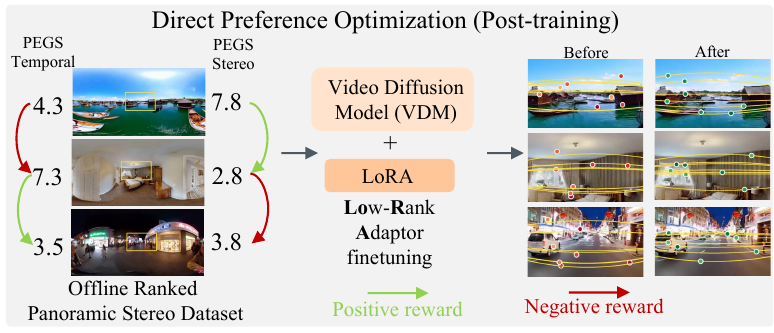}
    \caption{\textit{DPO.} We finetune a pre-trained panoramic video model using low-rank neural adaptors over stereo preference signal obtained from \textit{(good, bad)} video pairs.}
    \label{fig:dpo}
\end{figure}

  For each conditioning $c$ we sample several videos and score them with \cref{eq:pegs}. PEGS is
  an error, so the lowest-scoring sample is the preferred one $x_0^{w}$ and the highest-scoring
  the dispreferred one $x_0^{l}$. Each is noised independently through \cref{eq:rf_interp}.
  Writing $\mathcal{V}^{w}_{\theta}$ for the velocity error on $x^{w}_t$ and
  $\mathcal{V}^{w}_{\mathrm{ref}}$ for the same under the frozen model, and likewise for $l$,
  \cref{eq:rlhf} becomes the Flow-DPO objective
  \begin{equation}
      \mathcal{L}_{\mathrm{DPO}} = -\,\mathbb{E}\left[
      \log \sigma\!\left( -\frac{\beta_t}{2}
      \left(
        \mathcal{V}^{w}_{\theta} - \mathcal{V}^{w}_{\mathrm{ref}}
      - \mathcal{V}^{l}_{\theta} + \mathcal{V}^{l}_{\mathrm{ref}}
      \right)\right)\right],
      \label{eq:flow_dpo}
  \end{equation}
  where $\beta_t = \beta(1-t^2)$ weights the penalty by noise level and $\sigma$ is the logistic
  function. The bracket is small when the model denoises $x_0^{w}$ better than the reference and
  $x_0^{l}$ worse, so minimizing it favours geometrically consistent samples (\cref{fig:dpo}).

\subsection{Stage 3: Viewing Refinement}
  \label{sec:refinement}
\paragraph{Omnidirectional stereo for headset playback.}

\begin{figure}[h]
    \centering
    \includegraphics[width=\linewidth]{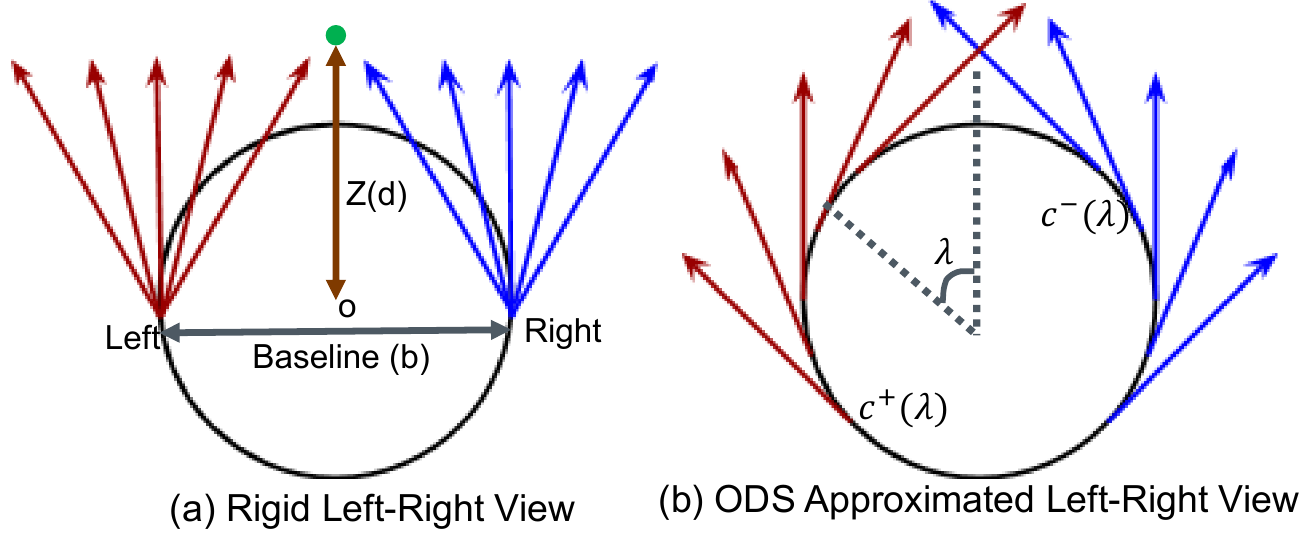}
    \caption{\textit{Rigid Stereo to ODS Correction.}}
    \label{fig:ods_viz}
\end{figure}

  Our pipeline places both eyes at fixed points in space, so the pair is only correct for a
  viewer facing one direction (\cref{fig:ods_viz}.a). Headset playback needs omnidirectional stereo, where the disparity
  is correct for every gaze direction (\cref{fig:ods_viz}.b).

  We convert it by re-sampling. Monocular depth has no scale of its own, so we first estimate depth for the left view and fix its scale by triangulating against the right view,
  \begin{equation}
      Z(\mathbf{d}) \;=\; \operatorname*{arg\,min}_{Z}
      \sum_{i} \bigl\lVert \pi_{R}\bigl(\mathbf{c}_{L} + Z(\mathbf{d}_i)\,\mathbf{d}_i\bigr)
      - \mathbf{m}_i \bigr\rVert^{2},
      \label{eq:ods_depth}
  \end{equation}
  where $Z$ is the depth along each ray, $\mathbf{c}_{L}$ the left-eye position, $\mathbf{d}_i$
  the left-view rays, $\mathbf{m}_i$ their matches in the right view and $\pi_R$ the right-eye
  projection. We then re-render, rotating the eye pair around a small circle and taking each
  output column from the eye position facing it,
  \begin{equation}
      \mathbf{c}^{\pm}(\lambda) \;=\; \mathbf{o} \;\pm\; \tfrac{b}{2}\,
      \hat{\mathbf{t}}(\lambda),
      \qquad
      \hat{\mathbf{t}}(\lambda) = (\cos\lambda,\; 0,\; \sin\lambda),
      \label{eq:ods_circle}
  \end{equation}
  where $\mathbf{c}^{\pm}$ are the two eye positions for the column at longitude $\lambda$,
  $\mathbf{o}$ is the rig midpoint and $b$ the interocular baseline.
  

  \paragraph{4K Upsampling.}
  Headset displays demand far more resolution than the generator produces, so we super-resolve
  both views before playback. Equi-rectangular frames cannot be fed directly to a stereo
  super-resolution network, so we split each view into cube faces and super-resolve the
  corresponding left--right face pairs with NAFSSR~\cite{nafssr}, then reassemble. Passing the two
  eyes through together keeps their detail consistent, which a per-view network would not
  guarantee. Ordering matters, as the omnidirectional stage re-renders both eyes from the left
  view, so we project first and super-resolve last.


\newcommand{\pending}{\textcolor{red}{$\ast$}}


\section{Results and Discussion}
\label{sec:results}

\subsection{Dataset and Implementation}
\label{sec:impl}

All generations use a frozen Wan2.1-1.3B backbone with the PanoWan \cite{panowan} equirectangular LoRA; no
backbone weights are updated. The left view is generated at $448\times896$ over $81$ frames
with $50$ denoising steps and guidance $5.0$. The right view is synthesized from it by
depth-guided backward warping at a rig baseline of $0.05$. Depth is generated on a per-frame basis using the panoramic monocular depth estimation model DAP \cite{dap}.

\begin{figure}[htbp]
    \centering
    \includegraphics[width=0.95\linewidth]{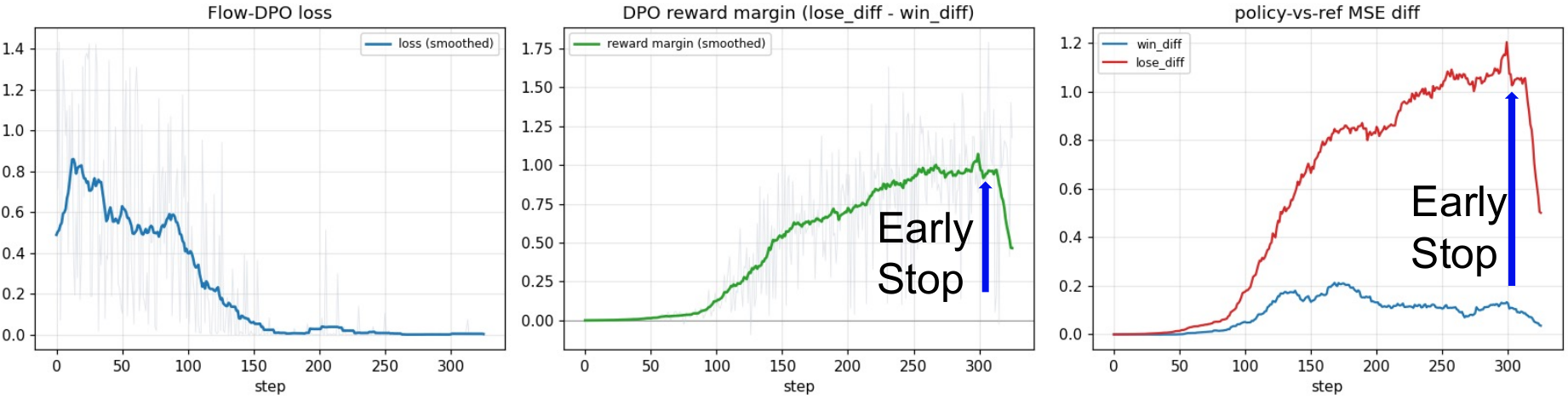}
    \caption{\textit{Flow-DPO training.} The reward margin creeps up while the policy's divergence from the
  reference accelerates, so we stop early rather than train to convergence.}
    \label{fig:refinement_results}
\end{figure}

We evaluate videos from two generation sets, \textbf{Stage 1} and \textbf{Stage 2}. For each stage we select $n=500$ videos, for evaluation on various axes as described next. For stage 2 refinement, we also generated an additional $n=500$ videos from the stage 1 zero-shot phase. We train a rank-$64$ LoRA ($\alpha = 128$) on the DiT attention and feed-forward projections,
  leaving the backbone and the PanoWan adapter frozen, with Adam at learning rate $5\times10^{-6}$
  and an effective batch of $4$. Preference pairs are formed only where the two samples differ by
  at least $5$\,mrad of PEGS, since pairs closer than that carry no reliable ordering. We check
  held-out PEGS every $25$ steps and stop early: \cref{fig:refinement_results} shows the reward margin
  creeping up while the policy's divergence from the reference accelerates, so late checkpoints
  drift from the reference model rather than improving on it.

\subsection{Metrics}
\label{sec:metrics}

We evaluate PEGS on three axes: \textbf{Stereo} compares left and right at fixed time. \textbf{Temporal} compares one
eye across time, under pose obtained from RANSAC with moving content emission. \textbf{Diagonal} compares one eye against the other
at a time offset. Lower is better on all three (measured in milli-radians on unit rays). A clip contributes to an axis only where a two-view pose is recoverable, which fails when parallax is too small to constrain it. We compare against Met3R~\cite{met3r}, a learned measure of multi-view consistency. It scores
a pair of views in feature space rather than as an angular residual, so it is unitless and defined only on perspective images. Both metrics see the same frames.

\paragraph{Properties.}
We identify seven core properties that our metric must follow:

\begin{itemize}\setlength{\itemsep}{0pt}\setlength{\parskip}{0pt}
  \item \textbf{Determinism:} The same input gives the same score.
  \item \textbf{Zero floor:} An exact pair scores zero.
  \item \textbf{Exchange symmetry:} Swapping left and right gives the same score.
  \item \textbf{Monotonicity:} A larger error never scores lower.
  \item \textbf{Frame invariance:} Turning the whole 360$^\circ$ stereo left-right rig does not change the score.
  \item \textbf{Calibration:} The score equals the error injected.
  \item \textbf{Motion robustness:} Moving objects are excluded, not charged as error.
\end{itemize}

\begin{table}[htbp]
  \centering
  \footnotesize
  \caption{Metric properties. $\checkmark$ holds, $\times$ violated,
  ``---'' not expressible.}
  \label{tab:metric_props}
  \begin{tabular}{lcc}
    \toprule
    Property & PEGS & MEt3R \\
    \midrule
    Determinism             & $\checkmark$ & $\checkmark$ \\
    Zero floor              & $\checkmark$ & $\checkmark$ \\
    Exchange symmetry       & $\checkmark$ & $\checkmark$ \\
    Monotonicity            & $\checkmark$ & $\times$     \\
    Frame invariance        & $\checkmark$ & ---          \\
    Calibration             & $\checkmark$ & ---          \\
    Motion robustness & $\checkmark$ & $\times$ \\
    \bottomrule
  \end{tabular}
\end{table}

As shown in \cref{tab:metric_props}, the two agree on the first three and differ on monotonicity, which decides whether a
metric can order generations at all. 
\begin{wrapfigure}[12]{r}{0.5\columnwidth}
  \centering
  \includegraphics[width=\linewidth]{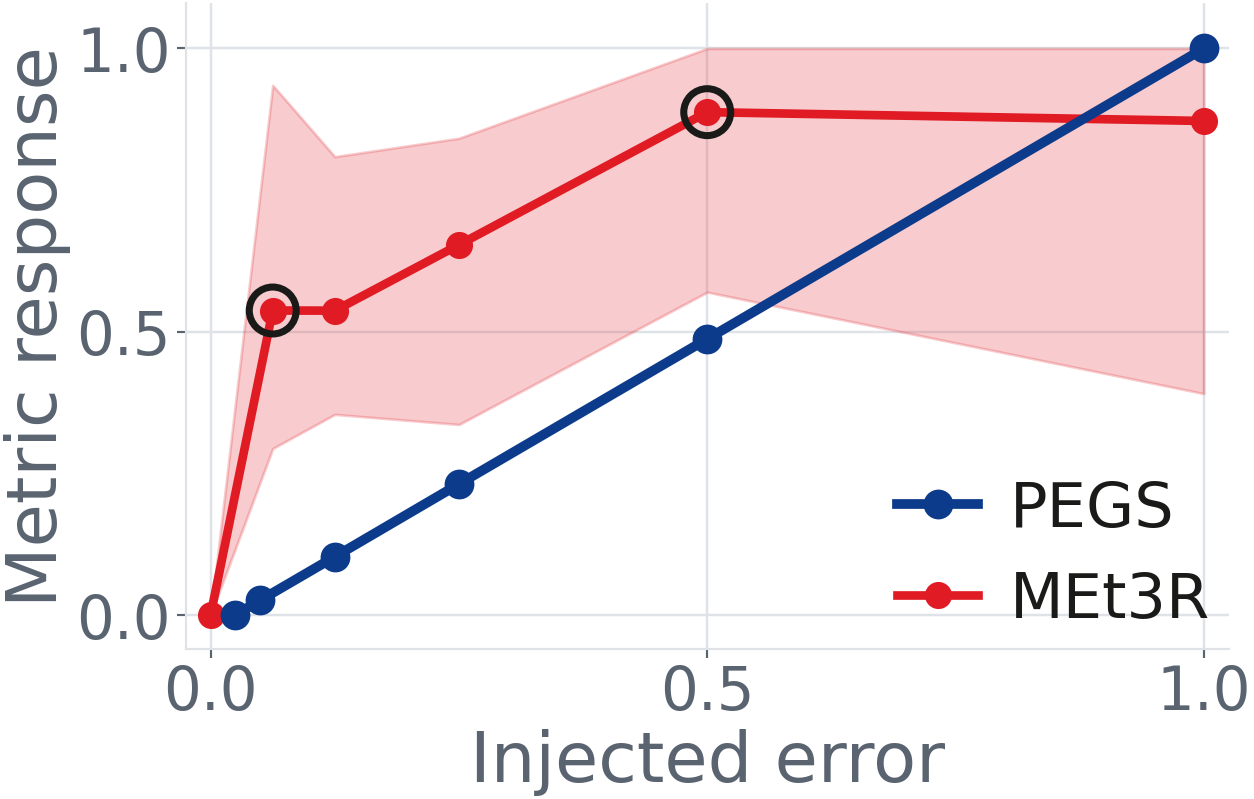}
  \caption{Response to injected misalignment.}
  \label{fig:metric_props}
  \vspace{-\intextsep}
\end{wrapfigure}
We start by injecting a frame misalignment of increasing size, and observe PEGS
increases strictly and returns exactly the error given, a slope of $1.000$ in milliradians;
MEt3R rises then falls on eight different sweeps (Fig.~\ref{fig:metric_props}). 
$0.38$--$0.68$\,mrad, MEt3R $-10\%$ to $+27\%$.

\begin{figure}[t]
  \centering
  \includegraphics[width=\columnwidth]{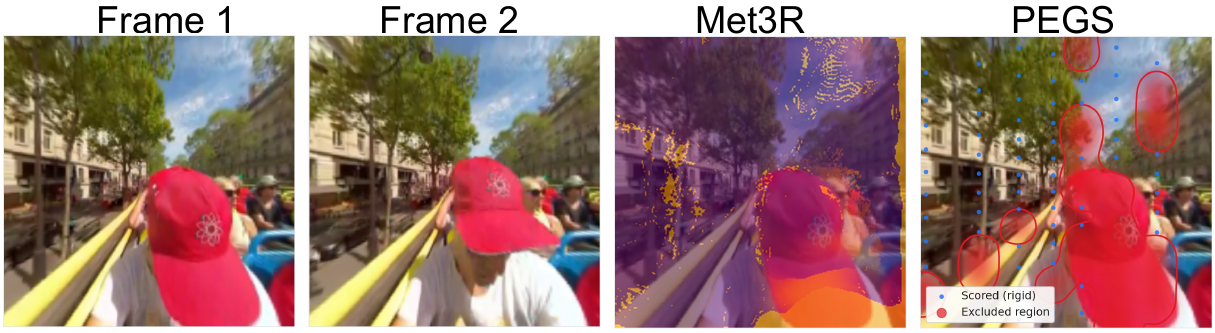}
  \caption{\textbf{Object motion breaks the rigid two-view model.} A passenger
  turns his head and torso between frame $t$ and $t{+}20$ while the background
  barely moves. MEt3R scores every pixel and charges the motion as
  inconsistency, whereas PEGS votes the moving tracks out (red, which also
  covers non-rigid foliage) and scores only the static scene: $1.10$ vs.\
  $2.54$\,mrad if they were retained.}
  \label{fig:motion_face}
\end{figure}

Frame invariance cannot be posed for MEt3R since by comparing feature maps, it never forms the rigid
two-view geometry needed for rotation. A feature distance carries
no angle, so it can rank pairs but never say by how much one is wrong. Calibration is hence not possible, given
$5$\,mrad of injected error PEGS returns $5$\,mrad, while MEt3R returns a feature distance corresponding to no physical quantity. Furthermore, under object motion our metric PEGS automatically eliminates motion while Met3R counts moving pixels towards matching breaking rigid two-view geometry assumptions as shown in \cref{fig:motion_face}.

\begin{figure*}[htbp]
    \centering
    \includegraphics[width=0.95\textwidth]{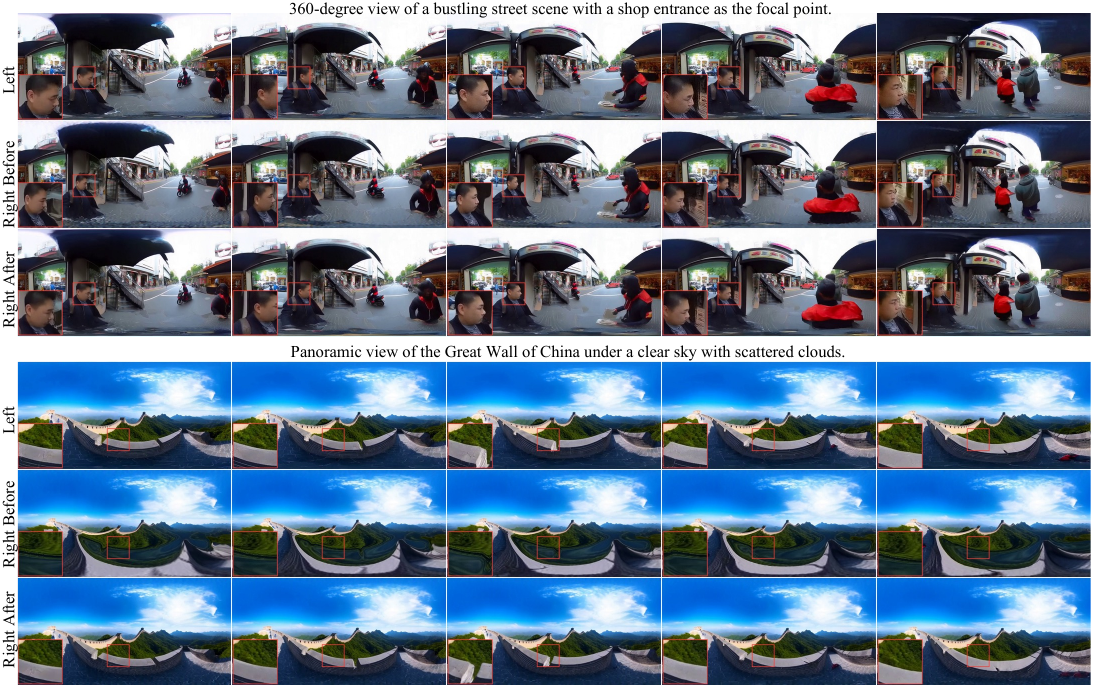}
    \caption{\textit{DPO refinement.} Our DPO refinement recovers fine geometry existing in the source left video lost during diffusion based inpainting over warped right view.}
    \label{fig:dpo_refine}
\end{figure*}

\subsection{Geometric Consistency}
\label{sec:pegs}

As shown in \cref{tab:pegs}, both metrics agree on the axis Stage~2 is meant to fix. Stereo consistency improves under PEGS
  ($3.121 \rightarrow 2.640$\,mrad) and under MEt3R ($0.096 \rightarrow 0.081$).
  \Cref{fig:dpo} shows the visual improvements in the output. Incorrect warp inpainting errors near the camera

  The two disagree on the temporal axis, and the disagreement is informative. This axis compares
  two frames of the same eye, so it responds to how much the scene moves as well as to how
  consistent it is. PEGS discards non-rigid tracks before scoring and reports an improvement
  ($1.494 \rightarrow 1.263$); MEt3R has no such test and reports the opposite. The diagonal axis
  asks for stereo and temporal consistency to hold at once and is the hardest to satisfy; neither
  metric shows a gain there, which is where the remaining headroom lies.

\begin{table}[htbp]
      \centering
      \footnotesize
      \caption{Evaluating PEGS and MEt3R on the two generation sets with $500$ videos each; lower is better.
      PEGS is a tangent-space angular residual in milliradians, MEt3R a unitless feature-space
      dissimilarity, so the two scales are not comparable to each other; only down each column.}
      \label{tab:pegs}
      \begin{tabular}{lcc cc}
        \toprule
        & \multicolumn{2}{c}{\textbf{PEGS} (mrad)}
        & \multicolumn{2}{c}{\textbf{MEt3R}~\cite{met3r} (unitless)} \\
        \cmidrule(lr){2-3}\cmidrule(lr){4-5}
        Axis & Stage 1 & Stage 2 & Stage 1 & Stage 2 \\
        \midrule
        Stereo    & $3.121$ & $\mathbf{2.640}$ & $0.096$ & $\mathbf{0.081}$ \\
        Temporal  & $1.494$ & $\mathbf{1.263}$ & $\mathbf{0.061}$ & $0.073$ \\
        Diagonal  & $2.958$ & $4.375$          & $0.104$ & $0.104$ \\
        \bottomrule
      \end{tabular}
    \end{table}

\begin{figure*}[htbp]
    \centering
    \includegraphics[width=0.95\textwidth]{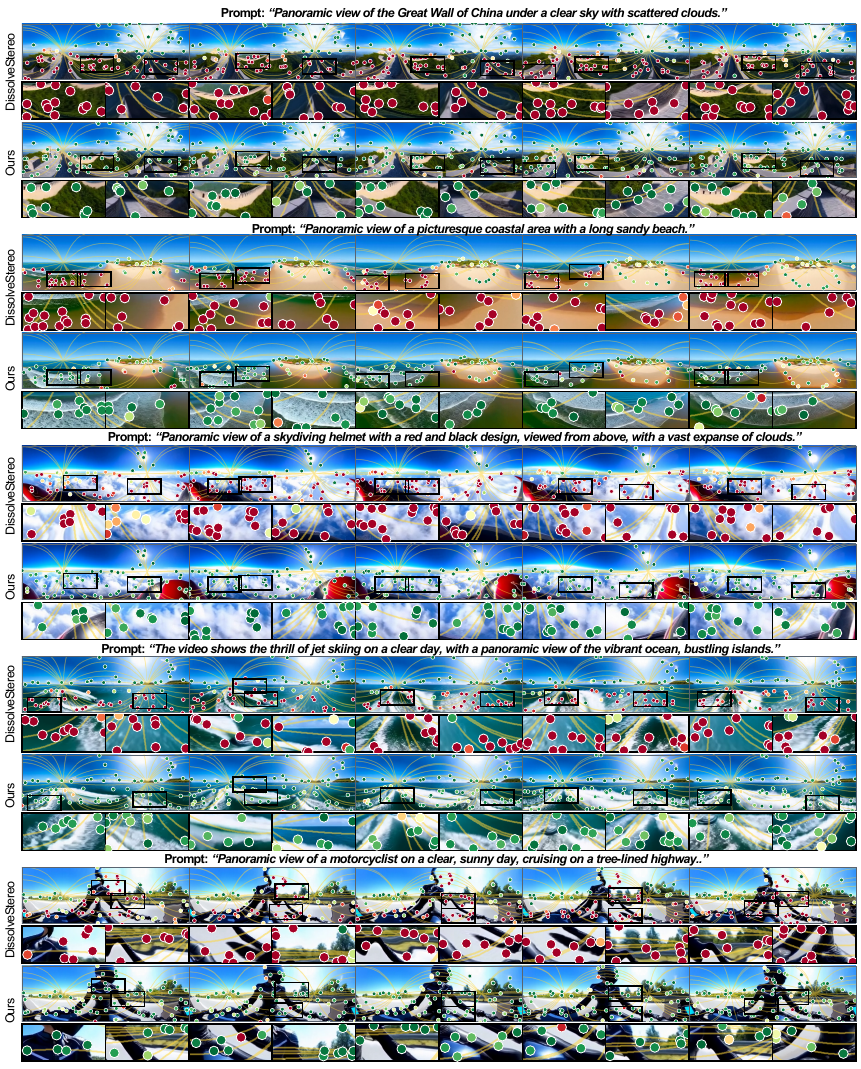}
    \caption{\textit{Comparing epipolar against DissolveStereo \cite{dissolvestereo}.} We evaluate DissolveStereo's zero-shot stereo adaptation procedure for panoramic stereo generation and compare against our method in terms of epipolar mismatches. \textcolor{red}{Red} indicates more epipolar mismatch and \textcolor{green}{Green} indicates less epipolar mismatches. }
    \label{fig:compare_prior}
\end{figure*}

\begin{figure*}[htbp]
    \centering
    \includegraphics[width=0.95\textwidth]{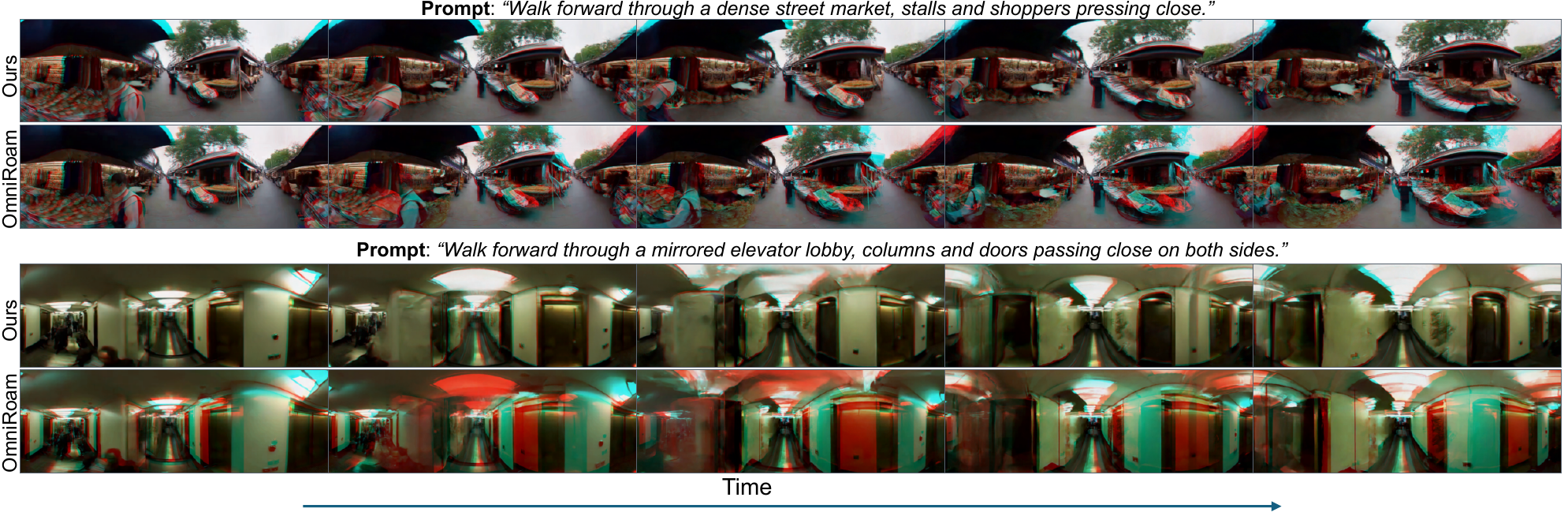}
    \caption{\textit{Camera Controlled Stereo [OmniRoam] vs Ours.} Our DPO based stereo adaptation performs better than adapting camera-conditioned video generation models such as OmniRoam which suffers from differing left-right generated views.}
    \label{fig:camera_stereo}
\end{figure*}
\

\subsection{Comparing against State of the Art}
\label{sec:sota}

DissolveStereo~\cite{dissolvestereo} generates stereo video from a monocular input by coarse
  depth injection with a noisy restart and iterative refinement, and is the closest published
  method to our right-view synthesis stage. It is a
  perspective method, with no mechanism for longitude wrap-around or the latitude-dependent
  sampling density of an equirectangular frame.

  Figure~\ref{fig:compare_prior} compares the two in terms of epipolar inconsistencies on the same
  sources (prompt $\rightarrow$ left view). We report this quantitatively in
  \cref{tab:epipolar}. The difference concentrates in the tail rather than the typical
  correspondence: the median residual barely separates the two methods, but correspondences that
  land far off their epipolar curve are three times rarer in our output. Ours is better on
  $46$ of the $50$ clips.

  We also compare against OmniRoam~\cite{liu2026omniroam}, a camera-controlled panoramic
  generator, qualitatively in \cref{fig:camera_stereo}. Because its camera can be placed
  anywhere, the obvious way to obtain a stereo pair is to generate the right eye a second time
  from a laterally shifted camera. Sharing the left eye between the two arms isolates that
  choice. \Cref{tab:omniroam} shows it does not produce a usable pair: the two eyes are each
  plausible but do not describe one scene. Ours is better on $9$ of the $10$ clips.

  \begin{table}[t]
    \centering
    \footnotesize
    \caption{Epipolar consistency against DissolveStereo, over $50$ clips sharing the same prompt,
    seed and left eye, so only the right eye differs. A correspondence is inconsistent when it
    lies $\ge 20$\,mrad off its epipolar curve. Lower is better.}
    \label{tab:epipolar}
    \begin{tabular}{lccc}
      \toprule
      Method & Inconsistent & Inconsistent (\%) & p95 (mrad) \\
      \midrule
      DissolveStereo & $282{,}949$         & $8.2$          & $15.4$ \\
      Ours           & $\mathbf{88{,}459}$ & $\mathbf{2.6}$ & $\mathbf{8.4}$ \\
      \bottomrule
    \end{tabular}
  \end{table}

  \begin{table}[t]
    \centering
    \footnotesize
    \caption{Against a camera pose-shift baseline, over $10$ clips sharing the same OmniRoam left
    eye, so only the right eye differs. Lower is better; disparity is reported to show the
    comparison is not won by a narrower baseline.}
    \label{tab:omniroam}
    \begin{tabular}{lccc}
      \toprule
      Method & Inconsistent (\%) & p95 (mrad) & Disparity (mrad) \\
      \midrule
      Pose-shift & $46.0$         & $132.9$        & $46.0$ \\
      Ours       & $\mathbf{1.2}$ & $\mathbf{8.2}$ & $21.5$ \\
      \bottomrule
    \end{tabular}
  \end{table}

\subsection{Analysis}
\label{sec:analysos}

\paragraph{Omnidirectional Stereo.}

\begin{figure}[t]
  \centering
  \includegraphics[width=\linewidth]{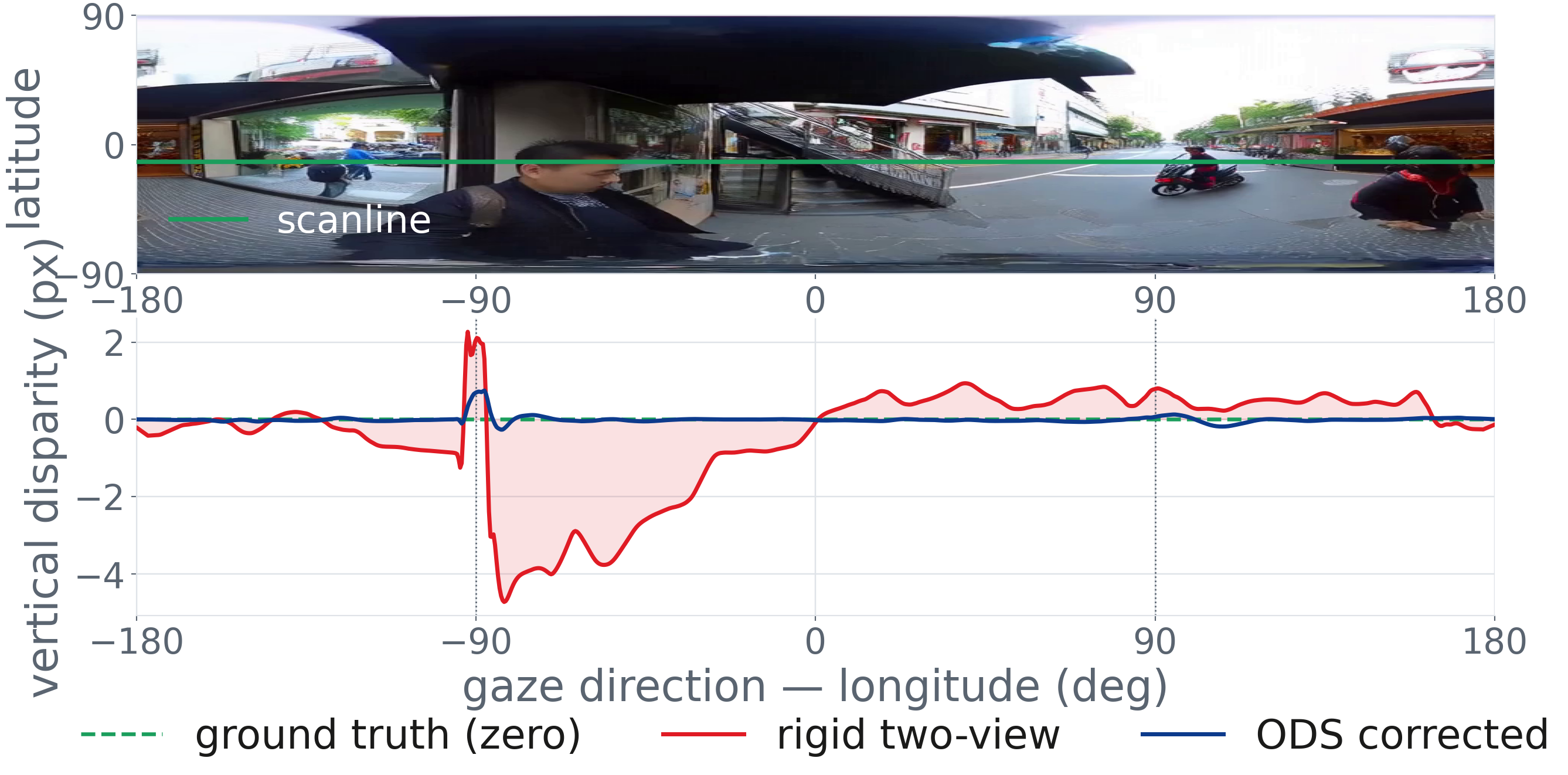}
  \caption{Vertical disparity along one scanline, measured between the delivered eyes. Zero is the ground truth for a fusable pair. The rigid rendering departs from it most where the baseline aligns with the gaze; the omnidirectional conversion holds it at zero throughout.}
  \label{fig:ods}
\end{figure}

  Our two eyes sit side by side, so a point should appear shifted only left or right between
  them, and that horizontal shift is what fuses into depth. A vertical shift cannot be fused at
  all, and simply causes strain. With a fixed rig the two eyes stop being side by side as soon as the viewer looks away from the forward direction, and that tilt appears as vertical shift. The ODS eye offset $\hat{\mathbf{t}}(\lambda)$ stays perpendicular to the gaze wherever the viewer
  looks, so no strain appears. \Cref{fig:ods} measures this on the delivered pixels: along one scanline
  the rigid pair reaches $4.73$\,px of vertical shift, worst near $\lambda = \pm 90^{\circ}$,
  while the converted pair stays at $0.04$\,px against a ground truth of exactly zero.

\vspace{-4pt}

\subsection{Ablations}
\label{sec:ablations}
\begin{figure}[htbp]
    \centering
    \includegraphics[width=\linewidth]{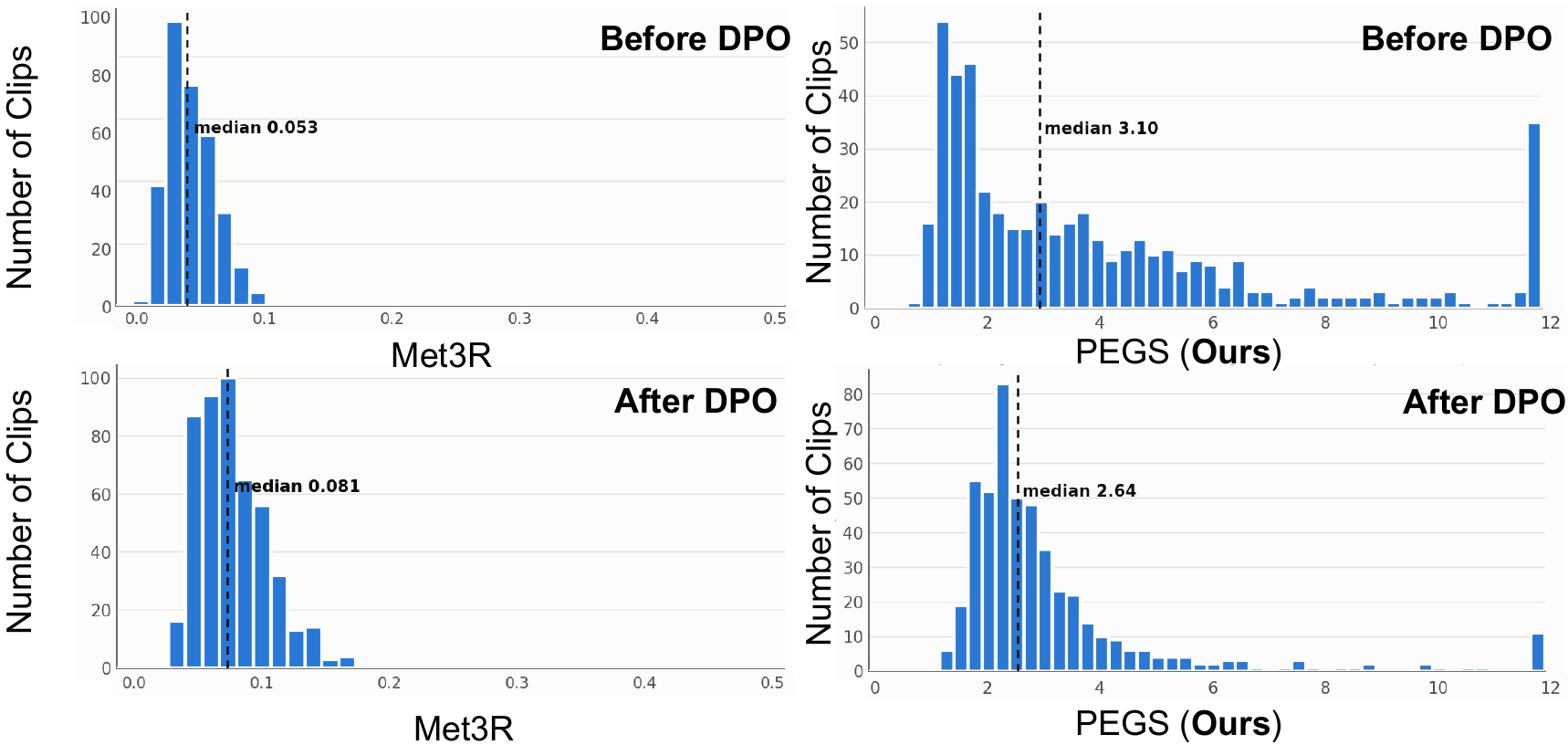}
    \caption{\textit{Met3R vs PEGS.} We perform DPO finetune ablation on both PEGS and Met3R with our framework and interestingly observe the \textit{tail-end} distribution improvement offered by PEGS while Met3R barely shows any improvement.}
    \label{fig:pegs_met3r_dpo}
\end{figure}

\paragraph{Impact of DPO Refinement:} We ablate the choice of preference signal by running the same DPO procedure with PEGS and with
  MEt3R as the reward (\cref{fig:pegs_met3r_dpo}). With PEGS the median improves from $3.10$ to
  $2.64$\,mrad, but the median understates what changed: the effect is concentrated in the tail,
  where clips scoring above $6$\,mrad thin out and the saturated bin at the top of the range falls
  from $35$ clips to $11$. This behavior is preferred from a geometric preference metric, since a
  clip whose stereo is wrong would be more noticeable than one that is marginally off.
  With MEt3R the distribution keeps its shape and its median moves the wrong way
  ($0.053 \rightarrow 0.081$). We attribute this to what each signal ranks: PEGS orders clips by
  measured epipolar violation, whereas MEt3R's feature-space score varies with scene content as
  much as with geometry, so the preference pairs it induces are not consistently ordered by
  consistency.







\vspace{-5pt}
\section{Conclusion}  We generate $360^\circ$ stereo video from a frozen monocular panoramic generator, with no stereo
  panoramic training data. The right eye comes from a spherical depth warp, and the model fills
  what the warp cannot. To measure the result we introduced PEGS, an angular epipolar residual on
  the viewing sphere that ignores non-rigid tracks and reports validity rather than a number when
  its assumptions fail. Used as the preference signal for DPO, it removes gross geometric failures
  rather than shifting the typical clip, and a final stage converts the pair to omnidirectional
  stereo for headset playback.

  PEGS assumes a rigid scene, so it can only score moving content by discarding it. The clearest
  next step is a constraint that ties all views to one scene at once: our hardest axis, which asks
  for stereo and temporal consistency together, is the one that does not yet improve.

\vspace{-\baselineskip}          









\bibliographystyle{abbrv-doi}

\bibliography{template}
\end{document}